\documentclass[10pt,twocolumn,letterpaper]{article}
\usepackage[pdftex]{graphicx}
\DeclareGraphicsExtensions{.pdf,.jpeg,.png}
\usepackage{cvpr}
\usepackage{times}
\usepackage{graphicx}
\usepackage{amsmath}
\usepackage{amssymb}
\usepackage{booktabs}
\usepackage{subfigure}
\usepackage{caption}
\usepackage[pagebackref=true,breaklinks=true,letterpaper=true,colorlinks,bookmarks=false]{hyperref}

\def\cvprPaperID{} 

\ifcvprfinal\fi
\begin{document}

\title{
Autonomous Deep Learning: A Genetic DCNN Designer for Image Classification
}

\author{Benteng Ma   \qquad      Yong Xia*\\
 School of Computer Science, Northwestern Polytechnical University\\
Xi’an 710072, China\\
{\tt yxia@nwpu.edu.cn}
}

\maketitle
\begin{abstract}
  Recent years have witnessed the breakthrough success of deep convolutional neural networks (DCNNs) in image classification and other vision applications. Although freeing users from the troublesome handcrafted feature extraction by providing a uniform feature extraction-classification framework, DCNNs still require a handcrafted design of their architectures. In this paper, we propose the genetic DCNN designer, an autonomous learning algorithm can generate a DCNN architecture automatically based on the data available for a specific image classification problem. We first partition a DCNN into multiple stacked meta convolutional blocks and fully connected blocks, each containing the operations of convolution, pooling, fully connection, batch normalization, activation and drop out, and thus convert the architecture into an integer vector. Then, we use refined evolutionary operations, including selection, mutation and crossover to evolve a population of DCNN architectures. Our results on the MNIST, Fashion-MNIST, EMNIST-Digit, EMNIST-Letter, CIFAR10 and CIFAR100 datasets suggest that the proposed genetic DCNN designer is able to produce automatically DCNN architectures, whose performance is comparable to, if not better than, that of state-of-the-art DCNN models.
\end{abstract}
\section{Introduction}
Deep convolutional neural networks (DCNNs), such as AlexNet ~\cite{krizhevsky01}, VGGNet ~\cite{simonyan02}, GoogLeNet ~\cite{szegedy03}, ResNet ~\cite{he04} and DenseNet ~\cite{huang05},  have significantly improved the baselines of most computer version tasks. Despite their distinct advantages over traditional approaches, DCNNs are still specialist systems that leverage a myriad amount of human expertise and data. They provide a uniform feature extraction-classification framework to free human from troublesome handcrafted feature extraction at the expense of handcrafted network design. Designing the architecture of DCNN automatically can not only bypass this issue but also take a fundamental step towards the long-standing ambition of artificial intelligence \ie. creating autonomous learning systems that require lest human intervention ~\cite{silver06}.

\begin{figure}
\begin{center}
\includegraphics[width=1.0\linewidth]{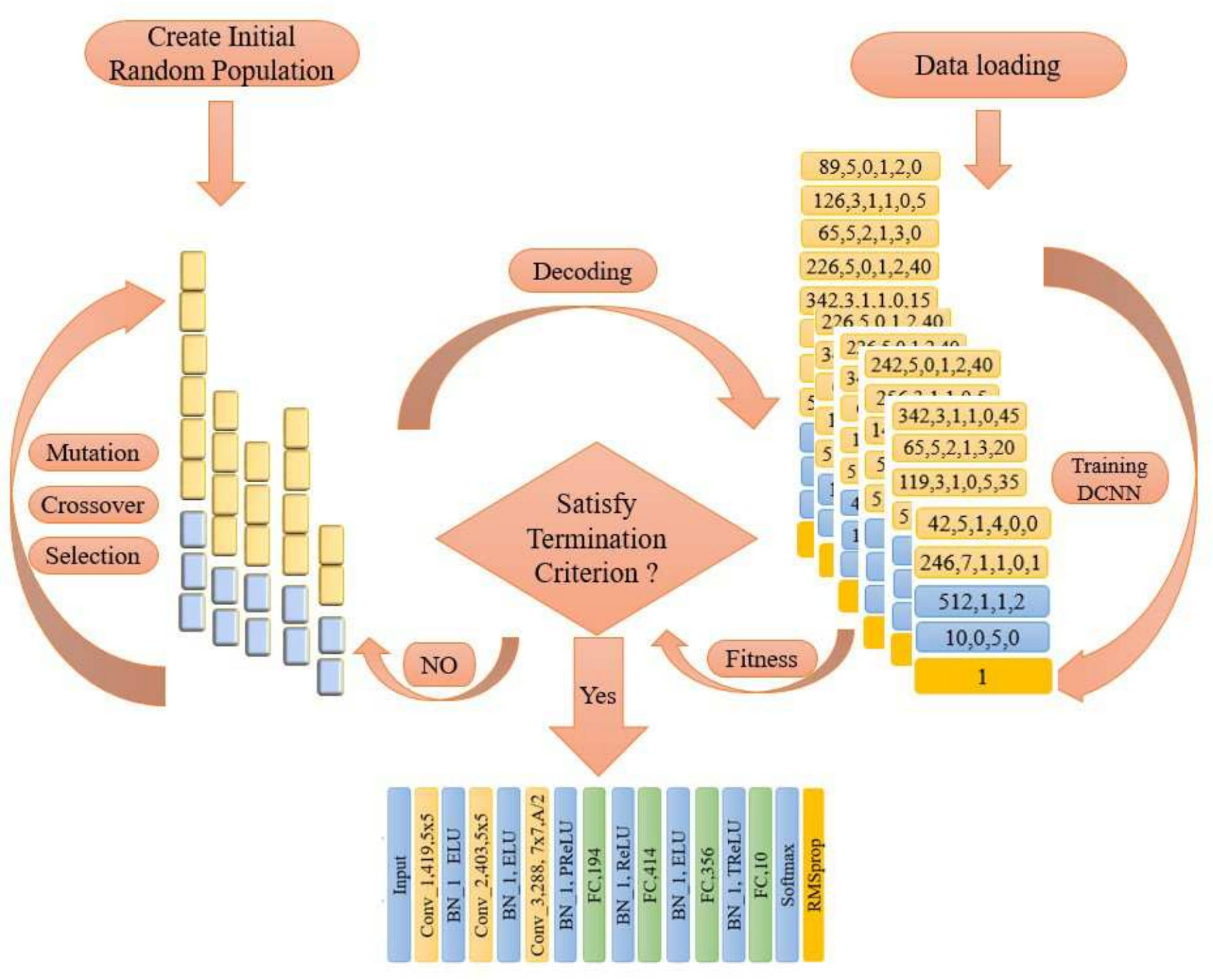}
\end{center}
   \caption{Diagram of the proposed genetic DCNN designer}
\label{fig:diagram}
\end{figure}
   Automated design of DCNN architectures has drawn more and more research attentions in recent years, resulting in a number of algorithms in the literature, which can be roughly divided into four categories: (1) DCNN architecture selection from a group of candidates, (2) DCNN architecture optimization using deep learning, (3) reinforcement learning-based DCNN architecture optimization, and (4) evolutionary optimization of DCNN architectures. Among them, evolutionary optimization approaches have a long history and seem to be very promising due to their multi-point global search ability, which enable them to quickly locate the areas of high quality solutions even in case of a very complex search space ~\cite{stanley07}. Despite their success, most evolutionary approaches pose restrictions either on the obtained DCNN architectures, such as the fixed depth, fixed filter size, fixed activation, fixed pooling operation ~\cite{xie08} and skipping out the pooling operation ~\cite{desell09}, or on the employed genetic operations, such as abandoning the crossover ~\cite{real10}. These restrictions may reduce the computational complexity, but also lead to lower performance. Alternatively, other evolutionary approaches may require hundreds even thousands of computers to perform parallel optimization ~\cite{desell09,real10}.
   
    In this paper, we propose a genetic DCNN designer to automatically generate the architecture of a DCNN for each given image classification problem. To reduce the complexity of representing a DCNN architecture, we develop a simple but effective encoding scheme, in which almost all the operations in a DCNN, such as the convolution, pooling, batch normalization, activation, fully connection, drop out and optimizer, are encoded as an integer vector. We start with a population of randomly initialized DCNN architectures and iteratively evolve this population on a generation-by-generation basis to create better architectures using the redefined genetic operations, including selection, crossover and mutation. We have evaluated our approach on six image classification tasks using the MNIST ~\cite{lecun11}, EMNIST-Digits ~\cite{cohen12}, EMNIST-Letters ~\cite{cohen12}, Fashion-MNIST ~\cite{xiao13}, CIFAR10 ~\cite{krizhevsky14} and CIFAR100 ~\cite{krizhevsky14} datasets. Our results indicate that the proposed genetic DCNN designer is able to generate automatically a DCNN architecture for each given image classification task, whose performance is comparable to the state of the art.
\section{Related Work}
  Many research efforts for automated design of DCNN architectures have been devoted to searching and selecting the most effective DCNN architecture from a group of candidates. Jin et al. ~\cite{jin15} introduced the sub-modular and super-modular optimization to the construction of DCNNs and established the guidelines to set the width and depth of DCNNs. Fernando \etal. ~\cite{fernando16} proposed the PathNet algorithm, which samples a sub-network from a super DCNN that has a much larger architecture, and demonstrated that this algorithm is capable of sustaining transfer learning on multiple tasks in both the supervised and reinforcement learning settings.

  Alternatively, the architecture and / or weights of a DCNN can be optimized by using another deep neural network. Ha \etal. ~\cite{ha17} used a hyper network and the discrete cosine transform (DCT) to evolve the weights of a DCNN, which has a fixed architecture. To generate the filters in a DCNN, Brabandere ~\cite{de18} proposed a dynamic filter network, which can be separated into a filter-generating network and a dynamic filtering layer. The former generates dynamically sample-specific filter parameters conditioned on the input, and the later applies those filters to the input.

  Recently, reinforcement learning has been applied to the design of DCNN architectures. Zoph and Le ~\cite{zoph19} employed a recurrent neural network trained by reinforcement learning to maximize the expected accuracy of the generated DCNN architecture on a validation set of images. This method, however, creates a DCNN with fixed-depth on a layer-by-layer basis, where each layer has a pre-determined number of filters and a pre-determined filter size. This method used distributed training and asynchronous parameters updated with 800 graphs processing units (GPUs). Barker \etal. ~\cite{baker20} proposed the MetaQNN method to generate DCNN architectures based on reinforcement learning. In this method, a Q-learning agent explores and exploits the space of model architectures with the $\epsilon-greedy$-strategy and experience replay.

  Various evolutionary algorithms have long been applied to the design of neural networks even before they became deep. Evolutionary optimization of neural network architectures can be traced back to NEAT, a seminal work done by Stanley and Miikkulainen ~\cite{stanley07}, which stores every neuron and connection in the DNA and generates architectures by mutation which can be divided into three kinds: (1) modifying a weight, (2) adding a connection between existing connections, and (3) inserting a neuron while splitting an existing connection. Based on this, Miikkulainen \etal. ~\cite{miikkulainen21} proposed the CoDeepNEAT algorithm, in which a population of chromosomes with minimal complexity is created and the structure is added to the graph incrementally through mutation over generations. Suganuma \etal. ~\cite{suganuma22} proposed an approach to design DCNN architecture based on genetic programming. The DCNN architecture is encoded by Cartesian genetic programming as directed acyclic graphs with a two-dimensional grid defined on the computational neurons.

  As one of the most prevalent evolutionary algorithms, the genetic algorithm (GA) ~\cite{Horn23} uses heuristics-guided search that simulates the process of natural selection and survival of the fittest. Xie \etal. ~\cite{xie08} encoded the architecture of each DCNN into a fixed-length binary string, and thus proposed a genetic DCNN (GDCNN) method to generate DCNN architectures automatically. In this method, the coding of DCNN architecture only contains the operation between pooling layers and the fully connected layers are not encoded. The DCNNs generated by this method, however, have a limited number of layers, fixed filter size and fixed filter number. Real \etal. ~\cite{real10} evolved DCNN architectures on the CIFAR-10 and CIFAR-100 datasets using GA with the DNA based coding scheme, which is same as that in NEAT. DCNN architectures evolve through mutation operations and the conflicts in filter sizes are handled by reshaping non-primary edges with zeroth order interpolation. All DCNN architectures generated by this method are fully trained. This method uses a distributed algorithm with more than 250 computers. Desell ~\cite{desell09} proposed the EXACT method, which can evolve DCNNs with  flexible connection and filter size based on GA with an asynchronous evolution strategy. The EXACT method used over 4,500 volunteered computers and trained over 120,000 DCNNs on the MNIST dataset.
\section{Method}
  The proposed genetic DCNN designer evolves and improves a population of individuals, each encoding an admissible DCNN architecture. The population is randomly initialized. The fitness of each individual is evaluated as the performance of the DCNN encoded by that individual in a specific image classification problem. Based on the current generation, a new generation is produced by performing a combination of redefined genetic operators, including selection, crossover and mutation, to improve the overall fitness of individuals. The evolution is performed iteratively on a generation-by-generation basis until meting a stopping criterion is fulfilled or the number of generations attains a pre-defined number.The diagram that summarizes this algorithm is shown in Figure~\ref{fig:diagram}.
  
\subsection{Encoding scheme}
  Our DCNN architecture encoding scheme is inspired by the representation of locus on a chromosome. A chromosome can be divided into two components: p-arm and q-arm.  The ordered list of loci known for a particular genome is called a gene map. Intuitively, various operations in a DCNN can be viewed as the loci on a chromosome, and thus the architecture of a DCNN can be encoded as a gene map, where convolutional blocks compose a convolutional arm and fully connected blocks compose a fully connected arm.

  A convolutional block contains five operations: convolution, batch normalization, pooling, activation and drop out. The convolutional operation has two parameters: the number of filters `N' and the size of filters `S'. The operations of batch normalization, pooling, activation and dropout are denoted by `B', `P', `A' and `D', respectively. Thus, a convolutional block contains $l_c = 6 $ loci in sequence and can be encoded as [\textit{NSPBAD}]. Similarly, a fully connected block contains $l_f = 4 $ loci in sequence and can be encoded as [\textit{NBAD}], where `N' is the number of neurons in a fully connected layer, `B' represents the operation of batch normalization, `A' represents the activation , `D' represents the operation of drop out. Besides, the optimizer also has an impact on the DCNN architecture, and is represented as `O'.
  
\begin{table}[htbp]
\small
\begin{center}
\caption{Value range of each at each locus of a network code}
\begin{tabular}{cc}
\hline 
  Code & Value scale \\
 \hline 
  N & [16, 512] \\
   S & { 3, 5, 7 } \\
  P & {0, 1, 2}\\
  B & {0, 1}\\
  A & {0, 1, 2, 3, 4}\\
  D & [0, 0.5]\\
  O & {0, 1, 2, 3, 4, 5, 6}\\
\hline 
\end{tabular} 
\end{center}
\end{table}
  Table 1 shows the value range at each locus of a network code: (1) the number of filters `N' ranges from 16 to 512; (2) the alternative sizes of filters are 3x3, 5x5 or 7x7; (3) the pooling operation P may take a value of 0 (without pooling), 1 (max pooling) or 2 (average pooling), all using the same stride of 2; (4) the batch normalization operation B takes a value from 0,1 to indicate whether adopting this operation or not; (5) the value of the activation `A' ranges from 0 to 5, representing the TReLU ~\cite{konda24}, ELU ~\cite{Djork25}, PReLU ~\cite{he26}, LeakyReLU ~\cite{maas27}, ReLU ~\cite{glorot28} and Softmax, respectively; (6) the dropout operation D takes a value from [0, 0.5] which indicates the probability of the randomly drop; and (7) the value of optimizer `O' ranges from 0 to 6, representing the SGD ~\cite{duchi30}, RMSprop ~\cite{zeiler31}, Adagrad ~\cite{zeiler31}, Adadelta ~\cite{kingma32}, Adam ~\cite{sutskever33}, Adamax ~\cite{sutskever33} and ~\cite{goldberg34}, respectively.
  
\begin{figure*}
\begin{center}
\includegraphics[width=0.8\linewidth]{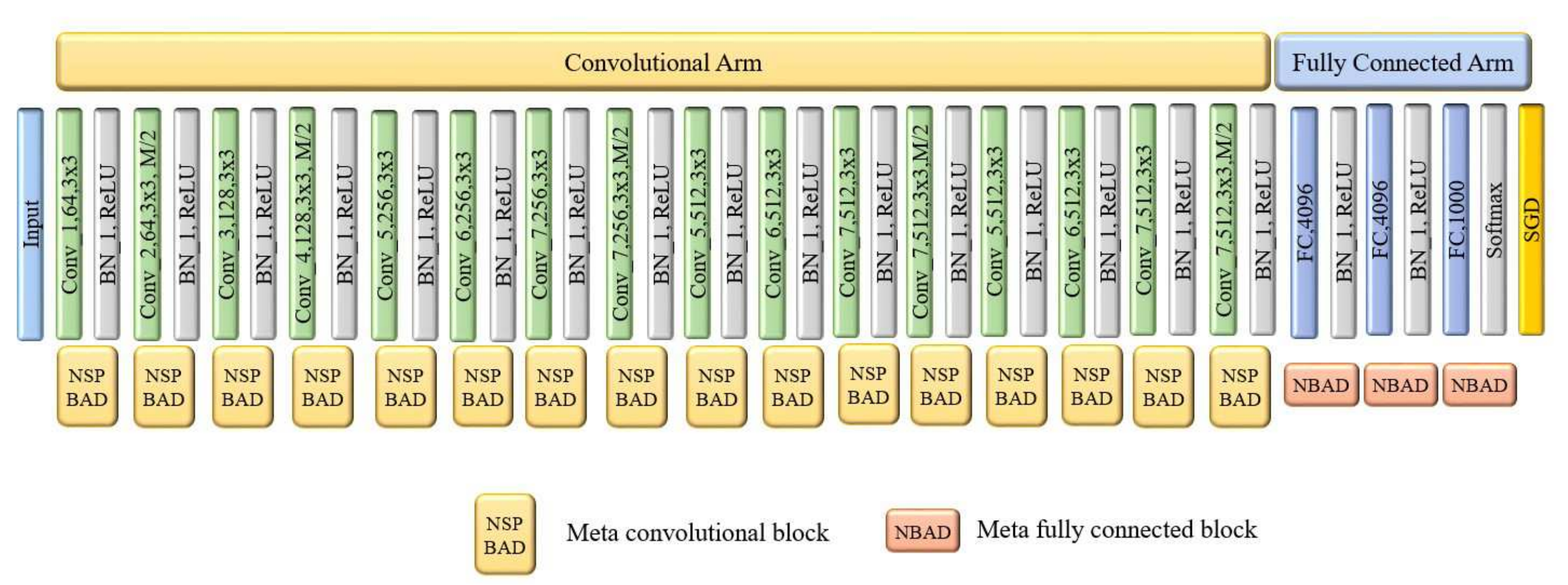}
\end{center}
   \caption{The blocks of VGGNet}
\label{fig:vgg}
\end{figure*}

With this coding scheme, a DCNN can be decomposed into a convolutional arm that contains a  sequence of convolutional blocks [NSPBAD] and a fully connected arm that contains a sequence of fully connected blocks [NBAD]. Taking the VGG-19 model shown in Figure~\ref{fig:vgg} as the case study, it can be presented as $\{ \{ [\textit{NSPBAD}]_{i}  \}^{16}_{i=1}, \{ [\textit{NBAD}]_{j}  \}^{3}_{j=1}, \textit{O} \} $ and be decoded as \{[64, 3, 0, 1, 4, 0], [64, 3, 2, 1, 4, 0], [128, 3, 0, 1, 4, 0], [128, 3, 2, 1, 4, 0], [256, 3, 0, 1, 4, 0], [256, 3, 0, 1, 4, 0], [256, 3, 0, 1, 4, 0], [256, 3, 2, 1, 4, 0], [512, 3, 0, 1, 4, 0], [512, 3, 2, 1, 4, 0], [512, 3, 0, 1, 4, 0], [512, 3, 2, 1, 4, 0], [512, 3, 0, 1, 4, 0], [512, 3, 2, 1, 4, 0], [512, 3, 0, 1, 4, 0], [512, 3, 2, 1, 4, 0]\}, \{[4096, 1, 4, 0], [4096, 1, 4, 0], [1000, 0, 5 , 0], 0 \}.
\subsection{Initialization}
  A DCNN with $ N_{n}^{C}$ convolutional blocks and $ N_{n}^{F}$ fully connected blocks can be presented as 
\begin{align}
S_{n} = \{ \{ [\textit{NSPBAD}]_{i}  \}^{N_{n}^{C}}_{i=1}, \{ [\textit{NBAD}]_{j}  \}^{N_{n}^{F}}_{j=1}, \textit{O} \}
\end{align}
whose code-length is
\begin{align}
L_{n}=N_{n}^{C}*l_c + N_{n}^{F}*l_f
\end{align}
At the initialization step, we set $ N_{n}^{C} \epsilon [1, 20] $ and $ N_{n}^{F} \epsilon [1, 3] $ and randomly sampled a population of DCNN architectures, denoted by $ \{ S_{n}\}^{T}_{n=1}$.
\subsection{Selection}
  Before producing the next generation, we evaluate each individual’s fitness. Based on the fitness ranking, we use the elitism roulette wheel selection scheme ~\cite{goldberg34} to select 0.1T of top ranking individuals (elitist) from the current generation to carry over to the next, unaltered, and  select 0.9T of individuals for subsequent genetic operations. This selection strategy guarantees that the highest fitness obtained by the GA will not decrease from one generation to the next. 
\subsection{Crossover}
For a pair of selected DCNNs $S_{i}$ and $ S_{j}$, we randomly locate a cross point on each of them, which breaks the DCNN architecture into two segments. By swapping the segments of those two DCNN architectures, two new DCNNs $S_{i}^{'}$ and $S_{j}^{'}$ are generated, whose depths may be different from the depths of their parents. 

If the cross point $k_{i}$ is located within the $ m_{i}-th$ convolutional blocks $[\textit{NSPBAD}]_{m_{i}}$ on the convolutional arm of the DCNN $S_{i}$, its location can be expressed as $(m_{i}-1)*l_{c}+x$. Then, we require that the other cross point   $k_{j}$ is also locates on the convolutional arm of the DCNN $ S_{j}$ and its location can be expressed as $ (m_{j}-1)*l_{c}+x$. After the crossover, the code-length of two newly generated DCNNs are
\begin{align}
\left\{
\begin{aligned}
L_{i}^{'} = L_{i}+(m_{i}-m_{j})*l_{c}\\
L_{j}^{'} = L_{j}+(m_{j}-m_{i})*l_{c} \\
\end{aligned}
\right.
\end{align}

\begin{figure}
\begin{center}
\includegraphics[width=0.8\linewidth]{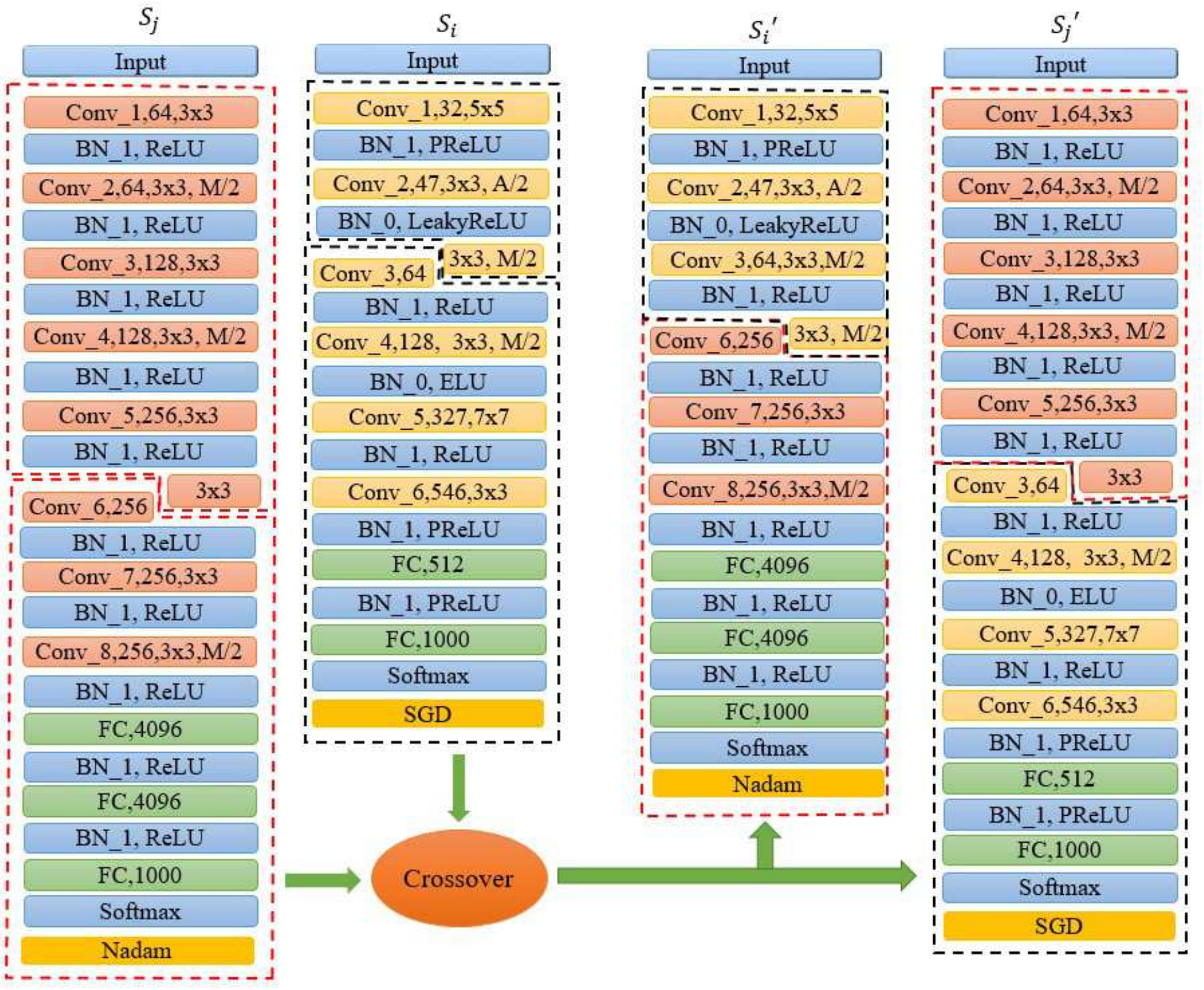}
   \caption{An example of crossover on the convolutional arm}
\label{fig:crossover}
\end{center}
\end{figure}

A typical  example is shown Figure~\ref{fig:crossover}, where one cross point $k_{i}$ is located at $3l_c+1$ in a DCNN with 8 learnable layers and the other cross point $k_{j}$ is located at $5l_c+1$  in a DCNN with 11 learnable layers. After the crossover, we obtained a DCNN with 9 learnable layers and a DCNN with 10 learnable layers.
If the cross point $k_{i}$ is located within the $ m_{i}-th$ fully connected blocks $[\textit{NBAD}]_{m_{i}}$ on the fully connected arm of the DCNN $S_{i}$, its location can be expressed as $ N_{n}^{C}* l_{c}+(m_{i}-1)*l_{c}+x$. Then, we require that the other cross point   $k_{j}$ also locates on the fully connected arm of the DCNN $ S_{j}$ and its location can be expressed as $ N_{n}^{C}* l_{c}+(m_{j}-1)*l_{c}+x$. After the crossover, the code-length of two newly generated DCNNs are
\begin{align}
\left\{
\begin{aligned}
L_{i}^{'} = L_{i}+(m_{i}-m_{j})*l_{f}\\
L_{j}^{'} = L_{j}+(m_{j}-m_{i})*l_{f} \\
\end{aligned}
\right.
\end{align}

\subsubsection{Mutation}
\begin{figure}
\begin{center}
\includegraphics[width=0.6\linewidth]{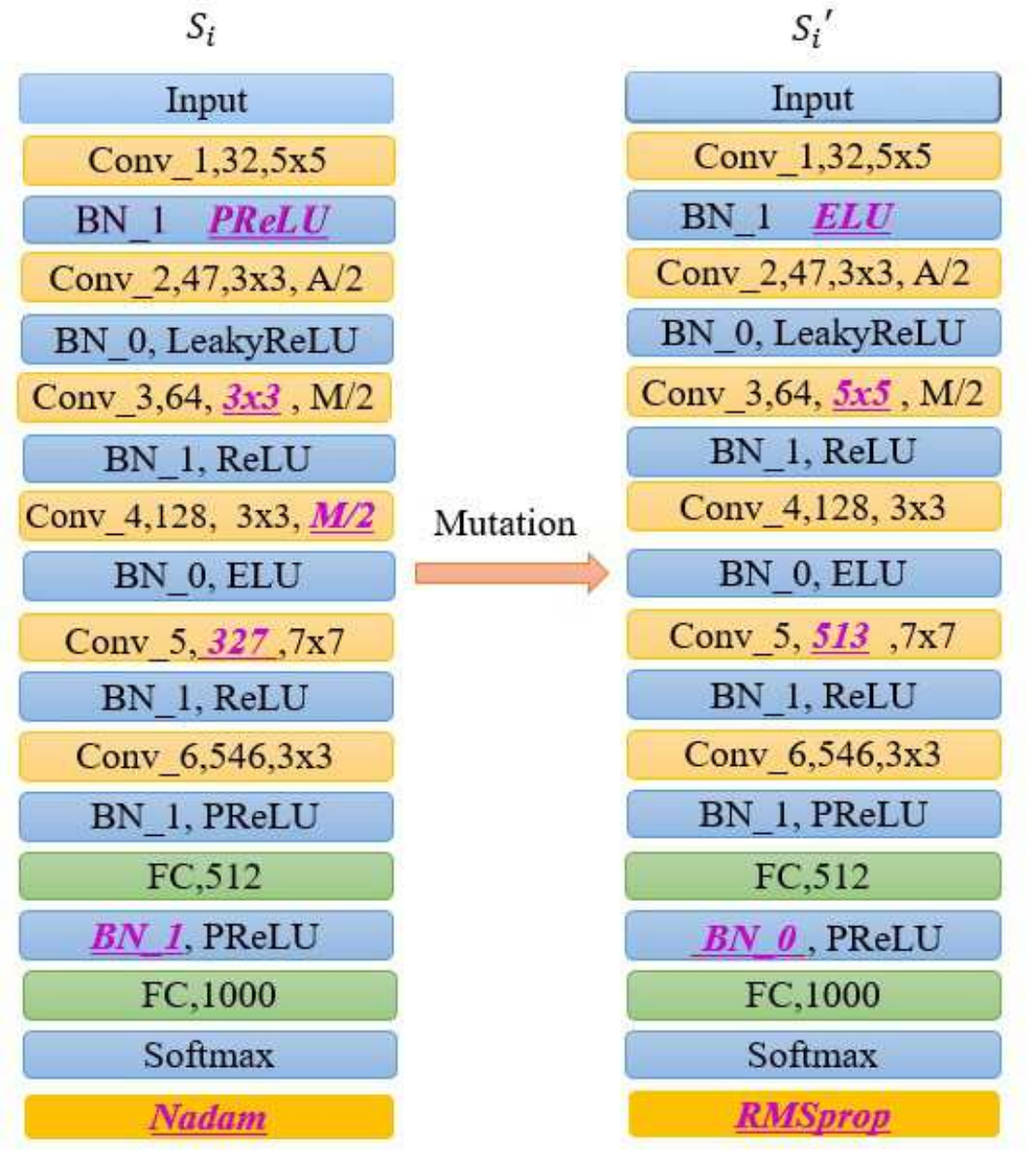}
   \caption{An example of mutation}
\label{fig:mutation}
\end{center}
\end{figure}

  To maintain genetic diversity from one generation to the next, the mutation operation is applied to each individual, which alters an DCNN architecture $ S_{n}$ by resampling each locus evenly and independently from the value range with a probability of $ q_{m}$. To accelerate the generation of new architectures, the value of $ q_{m}$ is evenly sampled from the range $ \left[ \dfrac{8}{L_{n}},0.5 \right]$ for each individual $ S_{n}$. An illustrative example of mutation is shown in Figure~\ref{fig:mutation}, where six loci in the code of $ S_{i}$ were mutated. After the mutation, the activation  in the first convolutional block was changed from PReLU to ELU, the kernel size in the third convolutional block was changed from 3x3 to 5x5, the max-pooling layer in the fourth convolutional block was removed, the number of kernels in the fifth convolutional block was changed from 327 to 513, the batch normalization in the first fully connected block was removed, and the optimizer was changed from Nadam to RMSprop. 
\section{Experiments and Results}
\subsection{Datasets}
  The datasets used for this study include the MNIST ~\cite{lecun11}, EMNIST-Digits ~\cite{cohen12}, EMNIST-Letters ~\cite{cohen12}, Fashion-MNIST ~\cite{xiao13}, CIFAR10 ~\cite{krizhevsky14} and CIFAR100 ~\cite{krizhevsky14}. The MNIST dataset defines a handwritten digit recognition task. The EMNIST-Letters and EMNIST-Digits datasets were derived from the NIST Special Database 19, which represents the final collection of handwritten characters, containing additional handwritten digits and an extensive collection of uppercase and lowercase handwritten letters. The EMNIST-Letters dataset merges all the uppercase and lowercase classes and gets a dataset with 26 classes comprising [a-z]. The CIFAR10 dataset is a subset of the 80-million tiny image database. In this dataset, both training and testing images are uniformly distributed over 10 categories. CIFAR100 is an extension to CIFAR10, and it contains 100 categories. Table 2 provides a brief summary of these six datasets, including the number of training images, testing images and classes, image size and color. 
\begin{table}
\scriptsize
\begin{center}
\caption{The summary of the datasets}
\begin{tabular}{cccccc}
\hline
& \# Training & \# Testing & Classes & size & Color \\
\hline
 MNIST & 60,000 & 10,000 & 10	& 28*28	& Grayscale \\
EMNIST-Letters & 124,800 & 20,800 & 26 & 28*28 & Grayscale \\
EMNIST-Digits & 240,000 & 40,000 & 10 & 28*28 & Grayscale\\
Fashion-MNIST & 60,000 & 10,000 & 10 & 28*28 & Grayscale\\
CIFAR10 & 50,000 & 10,000 & 10 & 32*32 & RGB\\
CIFAR100 & 50,000 & 10,000 & 100 & 32*32 & RGB\\
\hline
\end{tabular}
\end{center}
\end{table}
  For each dataset, the proposed genetic DCNN designer was used to generate a DCNN that can solve the corresponding image classification problem with a satisfying accuracy. In each experiment,In each experiment, 10\% of training images were used as a validation set, and others training images were used for training. To enlarge the training dataset, we applied the simple data augmentation method ~\cite{he04} to each training image. During the training of each DCNN, we use the cross-entropy loss  for all the DCNN architectures, fixed the maximum iteration number to 100, chose the min-batch stochastic gradient decent with a batch size of 256, set the learning rate as small as 0.0001 and further reduce it exponentially. 
\subsection{DCNN Designed for MNIST}
\begin{figure*}[htbp]
\centering                                                          
\subfigure[]{                   
 \begin{minipage}{2.5cm}
\centering
\label{fig:subfigure:a}
 \includegraphics[width=0.8\linewidth]{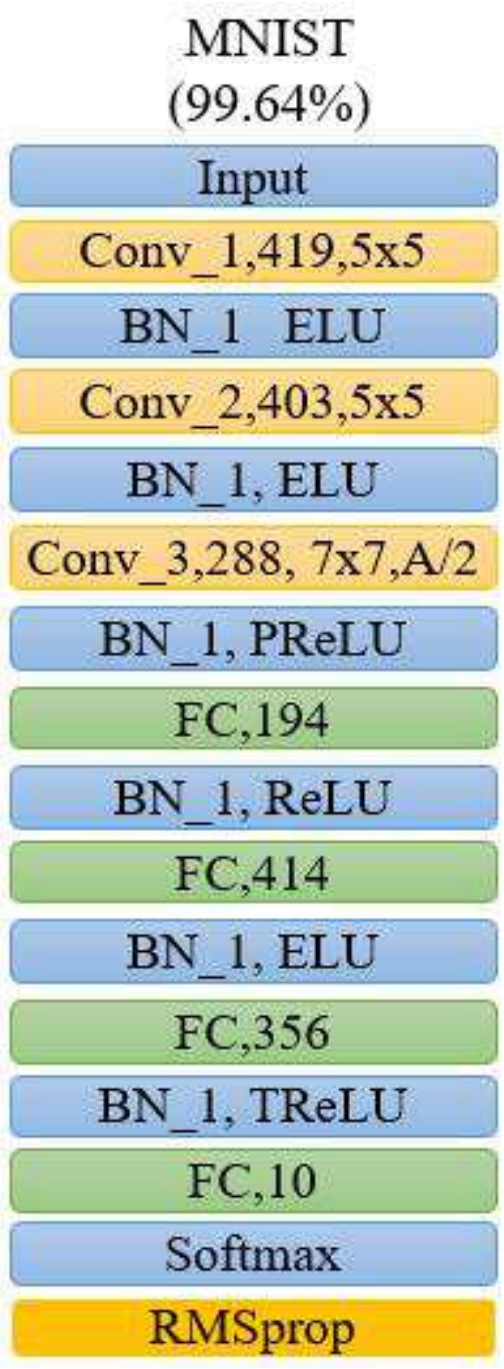}
               
  \end{minipage}
  }
\subfigure[]{                   
 \begin{minipage}{2.5cm}
\centering                                                         
\label{fig:subfigure:b}
 \includegraphics[width=0.8\linewidth]{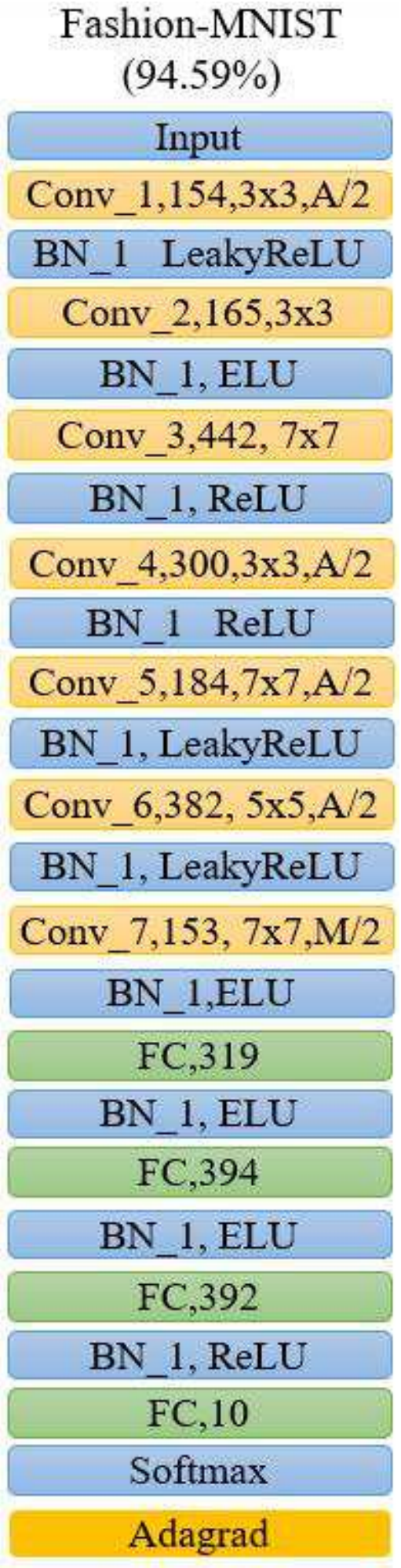}            
 \end{minipage}
}
\subfigure[]{                   
 \begin{minipage}{2.5cm}
\centering                                                         
\label{fig:subfigure:c}
 \includegraphics[width=0.8\linewidth]{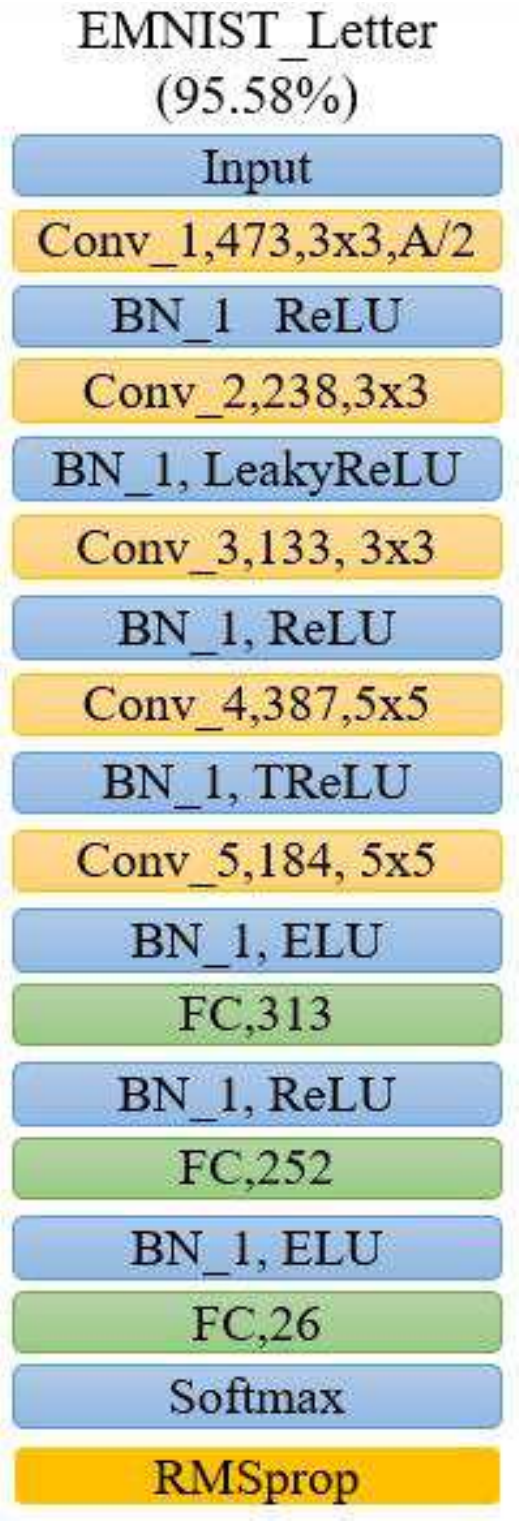}             
\end{minipage}
}
\subfigure[]{                   
 \begin{minipage}{2.5cm}
\centering                                                         
\label{fig:subfigure:d}
 \includegraphics[width=0.8\linewidth]{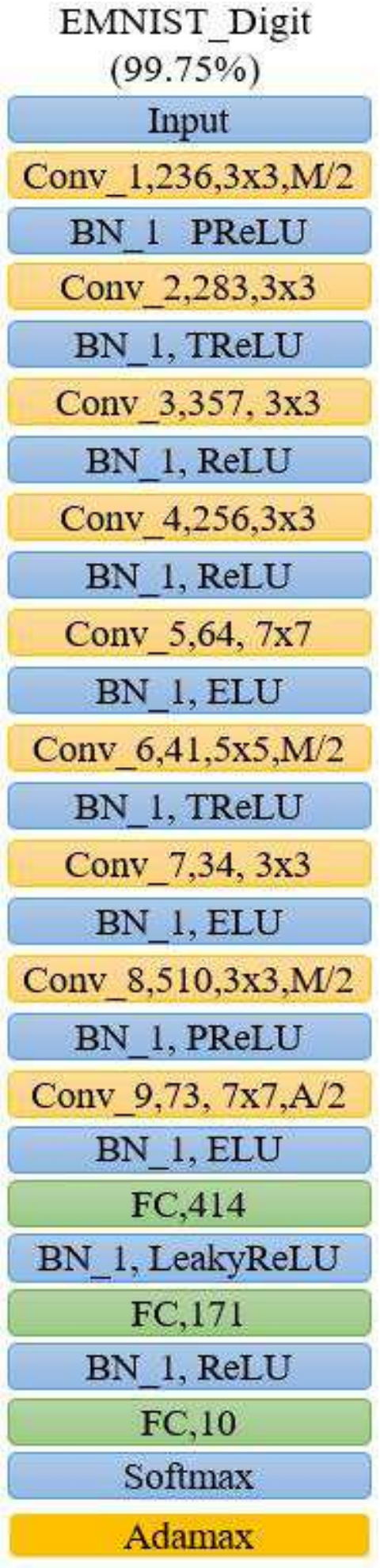}            
 \end{minipage}
}
\subfigure[]{                   
 \begin{minipage}{2.5cm}
\centering                                                         
\label{fig:subfigure:e}
 \includegraphics[width=0.8\linewidth]{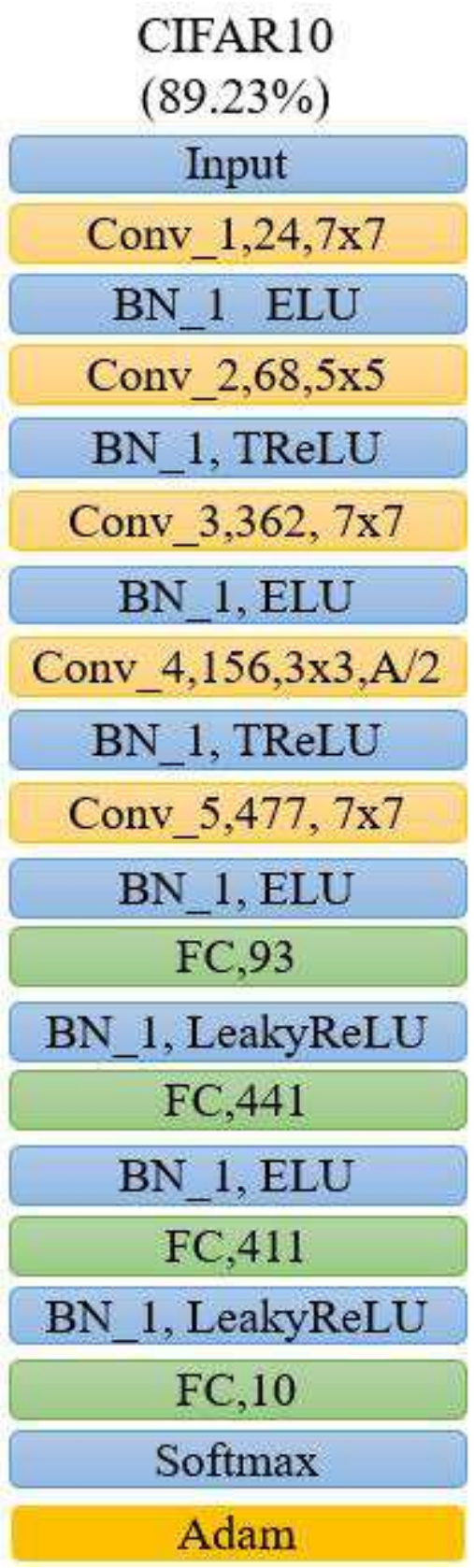}            
 \end{minipage}
}
\subfigure[]{                   
 \begin{minipage}{2.5cm}
\centering                                                         
\label{fig:subfigure:f}
 \includegraphics[width=0.8\linewidth]{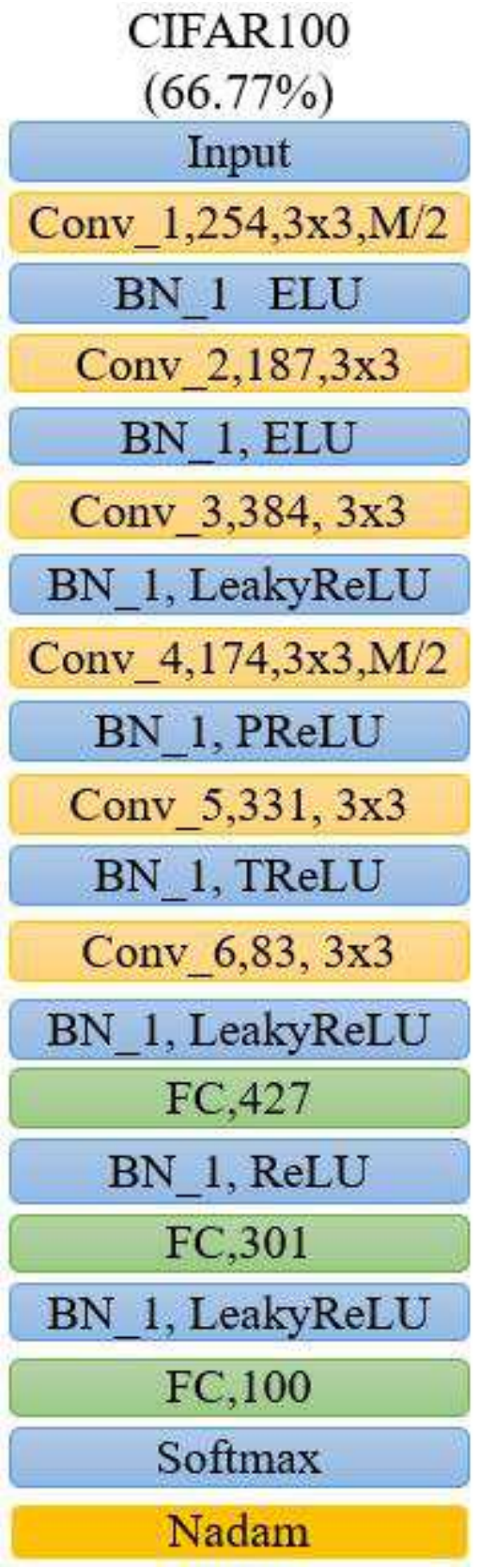}           
  \end{minipage}
}
\caption{The DCNN architectures generated by proposed genetic DCNN designer} \label{fig:subfigure}                                                        \end{figure*}

  The best DCNN architecture generated by the proposed genetic DCNN designer in ten generations for the MNIST dataset consists of three convolutional blocks and four fully connected blocks (see Figure~\ref{fig:subfigure:a} ). The first convolutional block contains 419 filters of size 5x5, no pooling layer, batch normalization, ELU activation  and 20\% dropout. The second convolutional block contains 403 filters of size 5x5, no pooling layer, batch normalization, ELU activation  and 0\% dropout. The third convolutional block contains 288 filters of size 7x7, an average-pooling  with stride of 2, batch normalization, PReLU activation  and 0\% dropout. The first fully connected block contains 194 neurons with the ReLU activation , batch normalization and 30\% dropout. The second fully connected block contains 414 neurons with the ELU activation , batch normalization and 45\% dropout. The third fully connected block contains 356 neurons with the TReLU activation , batch normalization and 5\% dropout. The last fully connected block is the output layer, which has 10 neurons with the softmax activation . The optimizer is Adamax.
\begin{table}
\footnotesize
\begin{center}
\caption{Classification accuracy of seven DCNNs on MNIST(\%)}
\begin{tabular}{cc}
\hline
 Method & Accuracy \\
\hline
AlexNet ~\cite{krizhevsky01} & 98.81 \\
VGGNet ~\cite{simonyan02} & 99.32 \\
ResNet ~\cite{he04}& 99.37\\
Capsule Net ~\cite{sabour35}  & 99.57\\
GDCNN ~\cite{xie08}& 99.65\\
EXACT ~\cite{desell09} & 98.32\\
\textbf{Proposed} & 99.72\\
\hline
\end{tabular}
\end{center}
\end{table}

  We compared the classification accuracy achieved by this architecture to the accuracy of AlexNet ~\cite{krizhevsky01}, VGGNet ~\cite{simonyan02}, ResNet ~\cite{he04} and Capsule Net ~\cite{sabour35} ] and the DCNNs generated by \textbf{GDCNN}~\cite{xie08} and EXACT ~\cite{desell09} in Table 3. It shows that the DCNN created by our genetic DCNN designer achieved the  highest accuracy of 99.72\%. The plot of highest classification accuracy achieved in each generation (see Figure~\ref{fig:acc:a}) shows that the performance of best DCNN architecture generated by our genetic DCNN designer has become stable.  
\subsection{DCNN Designed for Fashion-MNIST }
  The best DCNN architecture generated by the proposed genetic DCNN designer in ten generations for the Fashion-MNIST dataset consists of  seven convolutional blocks and four fully connected blocks (see Figure~\ref{fig:subfigure:b} ). The first convolutional block contains 154 filters of size 5x5, average pooling, batch normalization, LeakyReLU activation  and 0\% drop out. The second convolutional block contains 165 filters of size 3x3, no pooling , batch normalization, ELU activation  and 50\% drop out. The third convolutional block contains 442 filters of size 7x7, no pooling , batch normalization, ReLU activation  and 40\% drop out. The fourth convolutional block contains 300 filters of size 3x3, average pooling , batch normalization, ReLU activation  and 5\% drop out. The fifth convolutional block contains 184 filters of size 7x7, average pooling , batch normalization, LeakyReLU activation  and 25\% drop out. The sixth convolutional block contains 382 filters of size 5x5, average pooling , LeakyReLU activation  and 40\% drop out. The seventh convolutional block contains 153 filters of size 7x7, max pooling , batch normalization, ELU activation  and 40\% drop out. The first fully connected block contains 319 neurons with the ELU activation  and the 5\% drop out. The second fully connected block contains 394 neurons with the ELU activation  and 45\% drop out. The third fully connected block contains 392 neurons with ReLU activation  and 50\% drop out. The last fully connected block is the output layer, which has 10 neurons with the softmax activation . The optimizer is Adagrad. 
\begin{table}
 \tiny 
\begin{center}
\caption{Classification accuracy of four DCNNs on the four datasets(\%)}
\begin{tabular}{ccccc}
\hline
Method & MNIST &Fashion-MNIST & EMNIST-Letters & EMNIST-Digits\\
\hline
AlexNet ~\cite{krizhevsky01} & 98.81 & 86.43 & 89.36 & 99.21 \\
VGGNet ~\cite{simonyan02} & 99.32 & 90.45 & 94.62 & 99.62 \\
ResNet ~\cite{he04}& 99.37& 94.39 & 94.44 & 99.63\\
Capsule Net ~\cite{sabour35} & 99.57 & 90.03 & 91.58 & 99.37\\
Proposed & \bf{99.72 }& \textbf{94.60} & \textbf{95.58} & \textbf{99.75}\\
\hline
\end{tabular}
\end{center}
\end{table}

  The second column of Table 3 shows that, when compared to AlexNet ~\cite{krizhevsky01}, VGGNet ~\cite{simonyan02}, ResNet ResNet ~\cite{he04}, Capsule Net ~\cite{sabour35}, the architecture we generated achieved the highest accuracy of 94.60\%. The plot of highest classification accuracy obtained in each generation (see Figure~\ref{fig:acc:b}) shows that best DCNN architecture we created is still improving after 10 generations.
\subsection{DCNN Designed for EMNIST-Letters}
  The best DCNN architecture generated by the proposed genetic DCNN designer in ten generations for the EMNIST-Letters dataset  consists of five convolutional blocks and three fully connected blocks (see Figure~\ref{fig:subfigure:c}). The first convolutional block contains 473 filters of size 3x3, average pooling , batch normalization, ReLU activation  and 15\% drop out. The second convolutional block contains 238 filters of size 3x3, no pooling , batch normalization, LeakyReLU activation  and 20\% drop out. The third convolutional block contains 133 filters of size 3x3, no pooling , batch normalization, ReLU activation  and 10\% drop out. The fourth convolutional block contains 387 filters of size 3x3, no pooling , batch normalization, TReLU activation  and 10\% drop out. The fifth convolutional block contains 187 filters of size 5x5, no pooling , batch normalization, ELU activation  and 50\% drop out. The first fully connected block contains 313 neurons with ReLU activation , batch normalization and 20\% dropout. The second block contains 252 neurons with ELU activation , batch normalization and 20\% drop out. The last fully connected block is the output layer, which has 26 neurons with the softmax activation . The optimizer is RMSprop. 
  
  The third column of Table 3 shows that, when compared to AlexNet ~\cite{krizhevsky01}, VGGNet ~\cite{simonyan02}, ResNet ResNet ~\cite{he04}, Capsule Net ~\cite{sabour35}, the architecture we generated achieved the highest accuracy of 95.58\%. The plot of highest classification accuracy obtained in each generation (see Figure~\ref{fig:acc:c}) shows that best DCNN architecture we created is still improving after 10 generations.
\subsection{DCNN Designed for EMNIST-Digits}
  The best DCNN architecture generated by the proposed genetic DCNN designer in ten generations for the EMNIST-Digits dataset consists of nine convolutional blocks and three fully connected blocks (see Figure~\ref{fig:subfigure:d}). The first convolutional block contains 236 filters of size 3x3, max pooling , batch normalization, PReLU activation  and 0\% drop out. The second convolutional block contains 283 filters of size 5x5, no pooling layer, batch normalization, TReLU activation  and 30\% drop out. The third convolutional block contains 357 filters of size 3x3, no pooling , batch normalization, ReLU activation  and 35\% drop out. The fourth convolutional block contains 256 filters of size 3x3, no pooling , batch normalization, ReLU activation  and 15\% drop out. The fifth convolutional block contains 64 filters of size 7x7, no pooling , batch normalization, LeakyReLU activation  and 30\% drop out. The sixth convolutional block contains 41 filters of size 5x5, max pooling , batch normalization, TReLU activation  and 20\% drop out. The seventh convolutional block contains 34 filters of size 3x3, no pooling , batch normalization, ELU activation  and 25\% drop out is 0.25. The eighth convolutional block contains 510 filters of size 3x3, max pooling , batch normalization, PReLU activation  and 45\% drop out. The ninth convolutional block contains 73 filters of size 7x7, average pooling , batch normalization, ELU activation  and 5\% drop out. The first fully connected block contains 414 neurons with batch normalization, PReLU activation  and 20\% dropout. The second block contains 171 neurons with batch normalization, ReLU activation  and 40\% drop out. The last fully connected block is the output layer, which has 10 neurons with the softmax activation . The optimizer is Adamax.
  
  The fourth column of Table 3 shows that, when compared to AlexNet ~\cite{krizhevsky01}, VGGNet ~\cite{simonyan02}, ResNet ResNet ~\cite{he04}, Capsule Net ~\cite{sabour35}, the architecture we generated achieved the highest accuracy of 99.75\%. The plot of highest classification accuracy obtained in each generation (see Figure~\ref{fig:acc:d}) shows that best DCNN architecture we created has become stable after 10 generations.
\subsection{DCNN Designed for CIFAR10}
  The best DCNN architecture generated by the proposed genetic DCNN designer in ten generations for the CIFAR10 dataset consists of five convolutional blocks and four fully connected blocks (see Figure~\ref{fig:subfigure:e}). The first convolutional block contains 24 filters of size 7x7, no pooling , batch normalization, ELU activation  and 15\% drop out. The second convolutional block contains 68 filters of size 5x5, no pooling , TReLU activation  and 10\% drop out. The third convolutional block contains 362 filters of size 7x7, no pooling , batch normalization, ELU activation  and 0\% drop out. The fourth convolutional block contains 156 filters of size 3x3, average pooling, batch normalization, TReLU activation  and 0\% drop out. The fifth convolutional block contains 477 filters of size 7x7, average pooling, batch normalization, ELU activation  and 0\% drop out. The first fully connected block contains 93 neurons with batch normalization, LeakyReLU activation  and the 5\% of dropout. The second block contains 441 neurons with batch normalization, ELU activation  and 50\% drop out. The third block contains 411 neurons with batch normalization, LeakyReLU activation  and the 45\% drop out. The last fully connected block is the output , which has 10 neurons with the softmax activation . The optimizer is Adam.
\begin{table}
\small
\begin{center}
\caption{Classification accuracy on CIFAR10 and CIFAR100 (\%)}
\begin{tabular}{ccc}
\hline
 Method & CIFAR10 & CIFAR100 \\
\hline
AlexNet ~\cite{krizhevsky01} & 82.53 &60.53 \\
VGGNet ~\cite{simonyan02} & 84.62 &64.37 \\
ResNet ~\cite{he04}& \textbf{90.61} & 67.61\\
Highway Net  ~\cite{srivastava38} & 89.18 & 67.60\\
Capsule Net ~\cite{sabour35}& 89.40 & -\\
Proposed & 89.23 & \textbf{66.70}\\
\hline
\end{tabular}
\end{center}
\end{table}

\begin{figure}[htbp]
\small
\centering                                                          
\subfigure[]{                   
 \begin{minipage}{3.5cm}
\centering
\label{fig:acc:a}
 \includegraphics[width=1.2\linewidth]{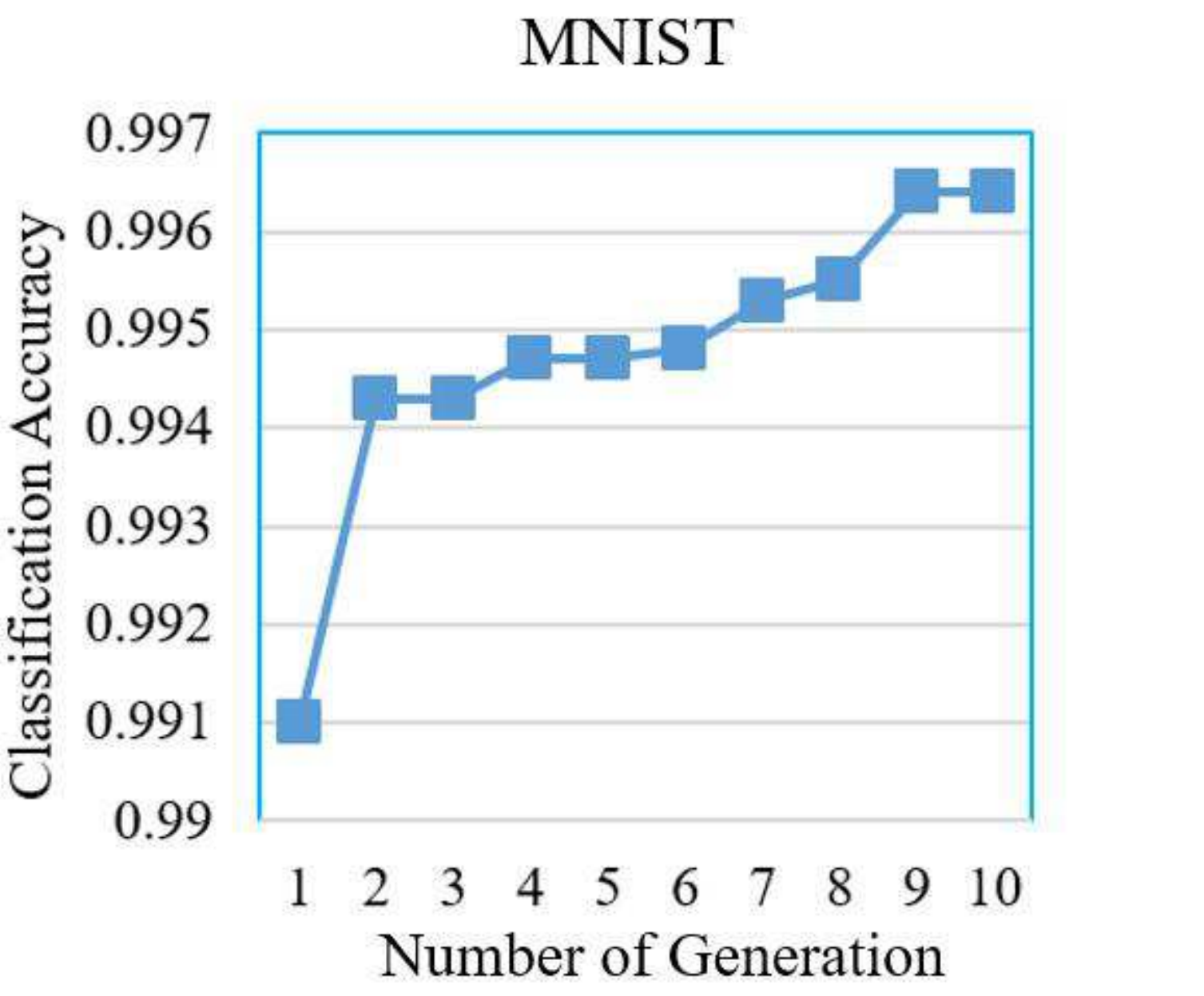}      
  \end{minipage}
  }
\subfigure[]{                   
 \begin{minipage}{3.5cm}
\centering                                                         
\label{fig:acc:b}
 \includegraphics[width=1.3\linewidth]{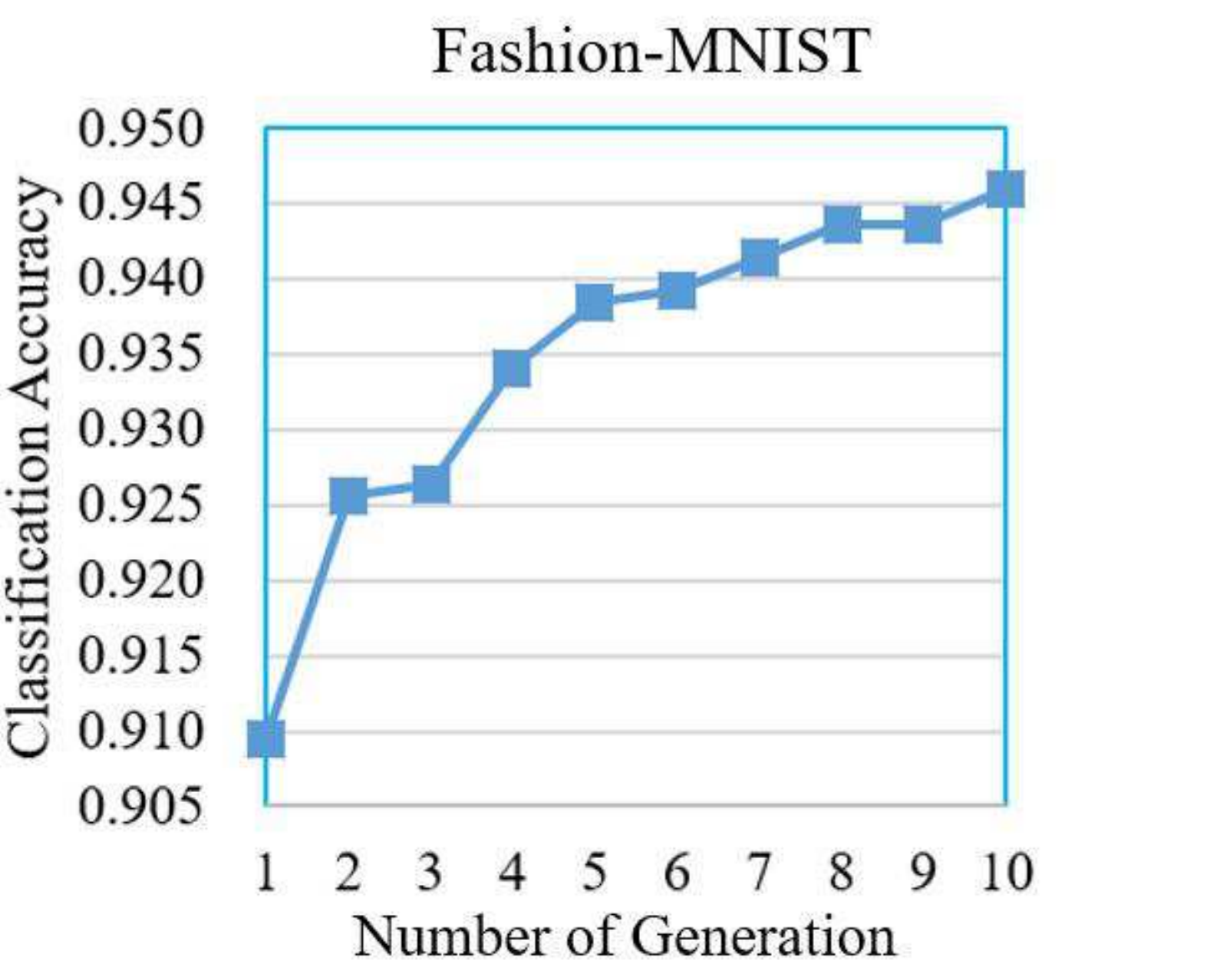}            
 \end{minipage}
}
\subfigure[]{                   
 \begin{minipage}{3.5cm}
\centering                                                         
\label{fig:acc:c}
 \includegraphics[width=1.2\linewidth]{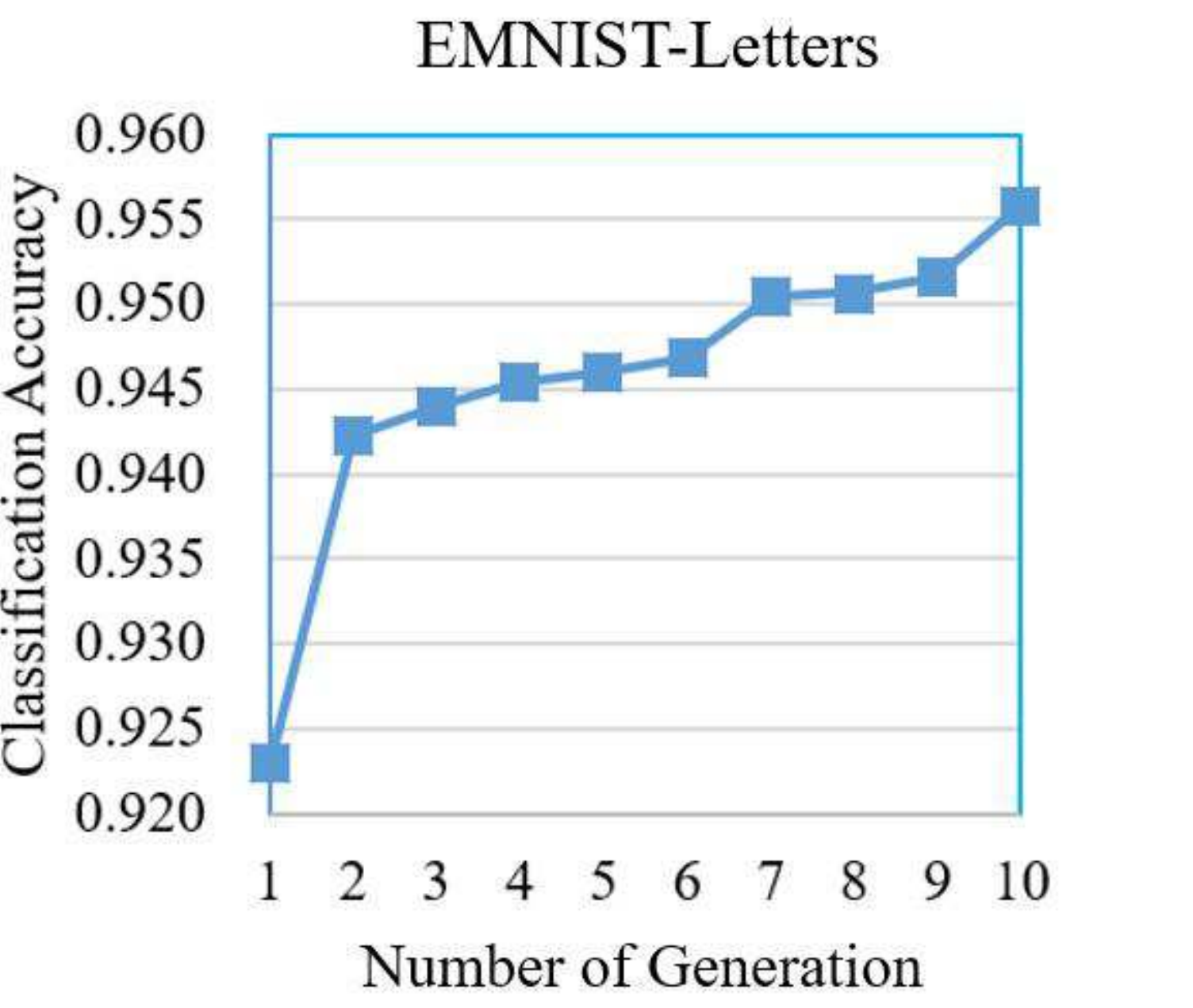}             
\end{minipage}
}
\subfigure[]{                   
 \begin{minipage}{3.5cm}
\centering                                                         
\label{fig:acc:d}
 \includegraphics[width=1.3\linewidth]{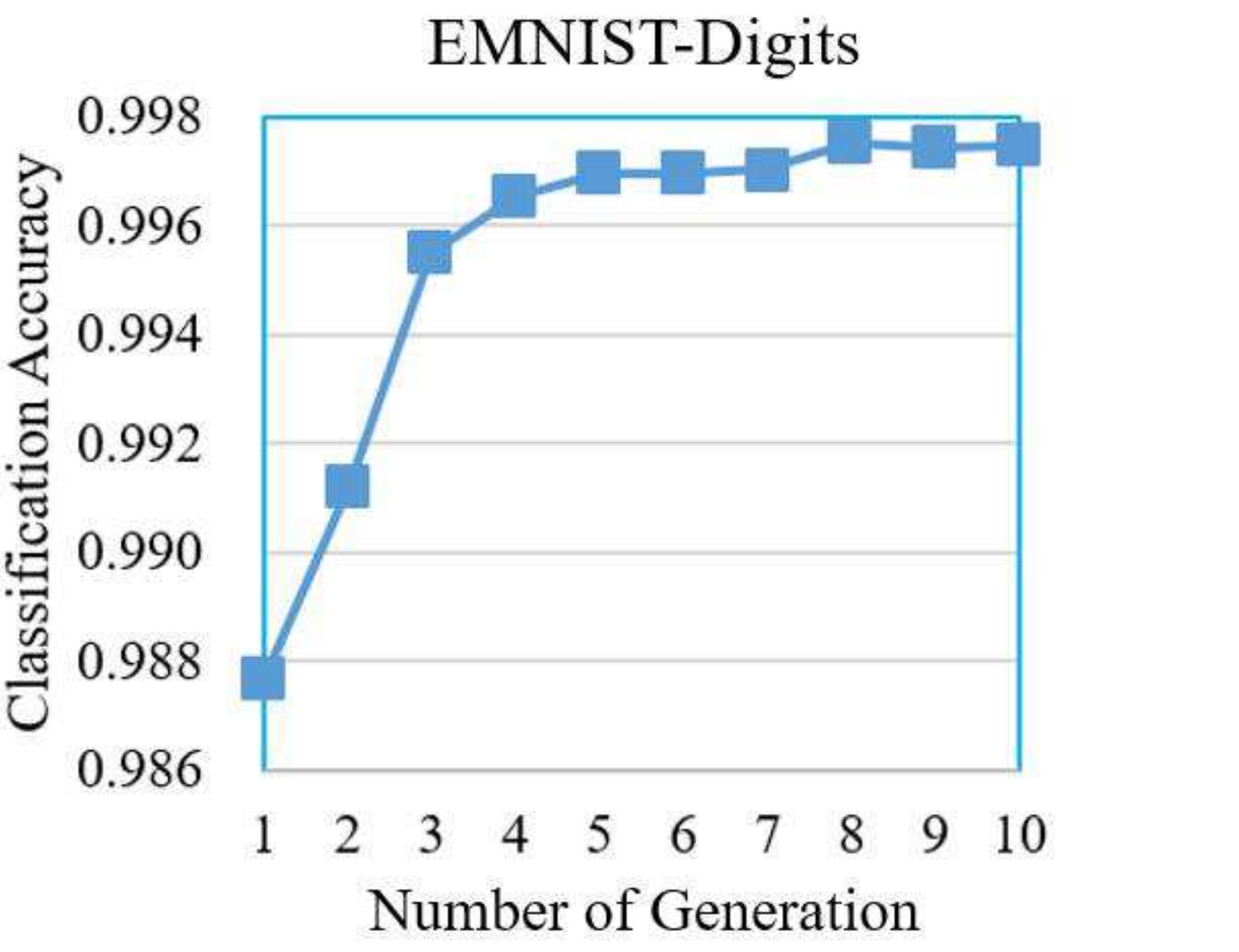}            
 \end{minipage}
}
\subfigure[]{                   
 \begin{minipage}{3.5cm}
\centering                                                         
\label{fig:acc:e}
 \includegraphics[width=1.2\linewidth]{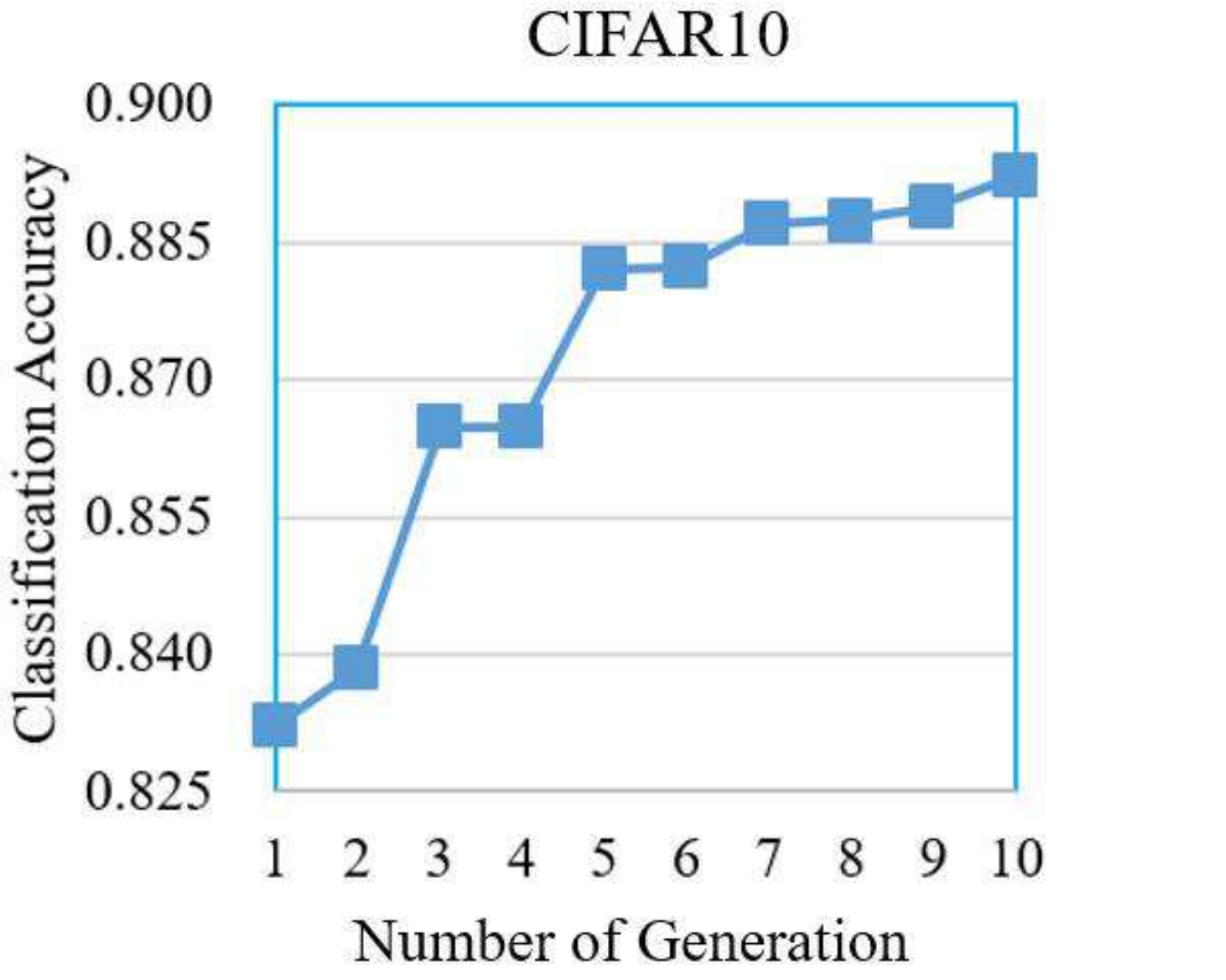}            
 \end{minipage}
}
\subfigure[]{                   
 \begin{minipage}{3.5cm}
\centering                                                         
\label{fig:acc:f}
 \includegraphics[width=1.2\linewidth]{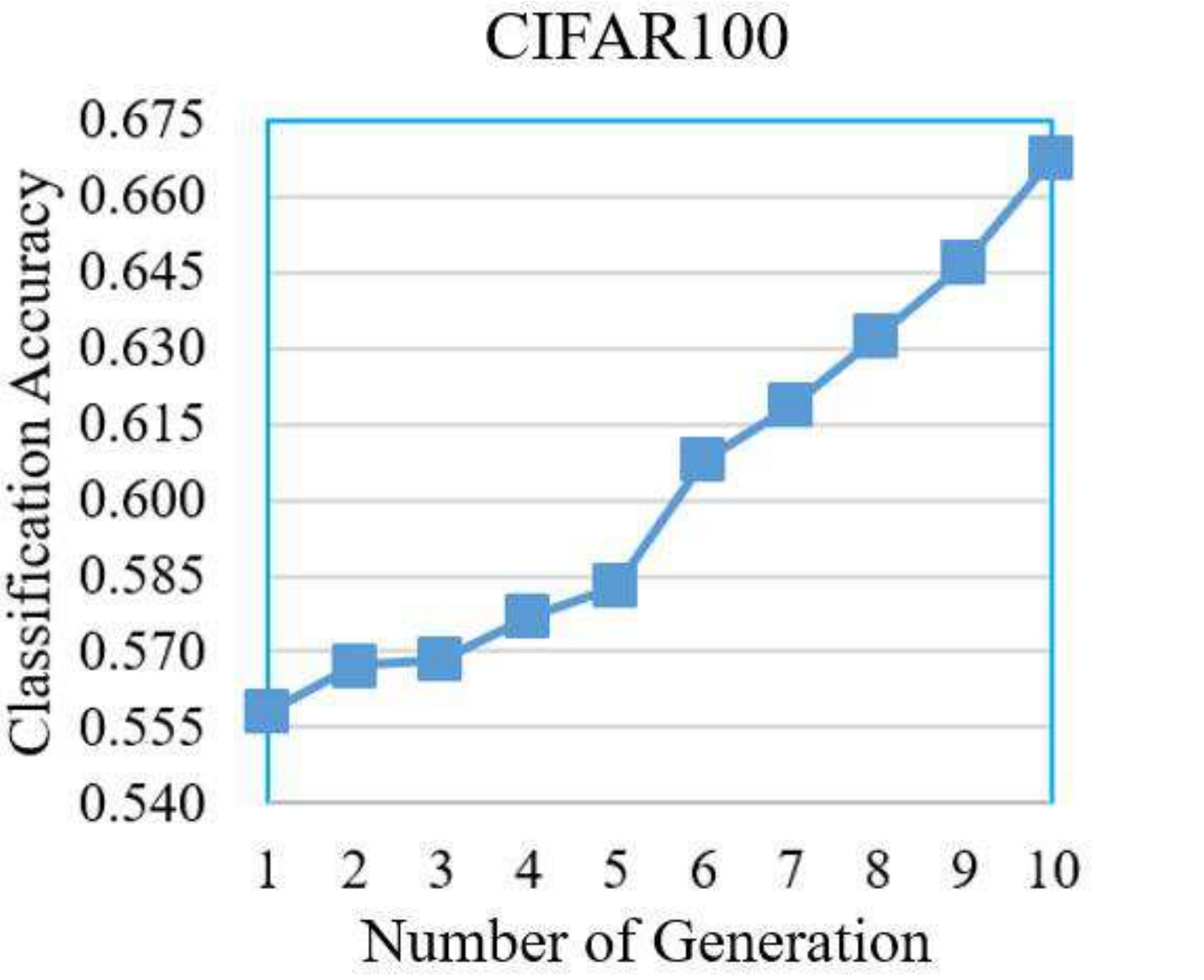}           
  \end{minipage}
}
\caption{Highest classification accuracy achieved in each generation on each dataset} \label{fig:acc}                                                        \end{figure}

  The second column of Table 4 shows that, when compared to AlexNet ~\cite{krizhevsky01}, VGGNet ~\cite{simonyan02}, ResNet ResNet ~\cite{he04},~\cite{srivastava38} and Capsule Net ~\cite{sabour35},, the architecture we generated achieved the third highest accuracy of 89.32\%, lower than the accuracy of Capsule Net (89.40\%) and ResNet (90.61\%). However, the plot of highest classification accuracy obtained in each generation (see Figure~\ref{fig:acc:e}) shows that the performance of best DCNN architecture we generated has become stable after 10 generations.

\subsection{DCNN Designed for CIFAR100}
  The best DCNN architecture generated by the proposed genetic DCNN designer in ten generations for the CIFAR100 dataset consists of six convolutional blocks and three fully connected blocks (see Figure~\ref{fig:subfigure:f}). The first convolutional block contains 254 filters of size 3x3, max pooling , batch normalization, ELU activation  and 0\% drop out. The second convolutional block contains 187 filters of size 3x3, no pooling , batch normalization, ELU activation  and 0\% drop out. The third convolutional block contains 384 filters of size 3x3, no pooling , batch normalization, LeakyReLU activation  and 30\% drop out. The fourth convolutional block contains 174 filters of size 3x3, max pooling, batch normalization, PReLU activation  and 20\% drop out. The fifth convolutional block contains 331 filters of size 3x3, no pooling , batch normalization, LeakyReLU activation  and 40\% drop out. The sixth convolutional block contains 83 filters of size 3x3, no pooling , batch normalization, LeakyReLU activation  and 35\% drop out. The first fully connected block contains 427 neurons, batch normalization, ReLU activation  and 35\% drop out. The second fully connected block contains 301 neurons with batch normalization, LeakyReLU activation  and 15\% drop out. The last fully connected block is the output layer, which has 10 neurons with the softmax activation . The optimizer is Nadam.
  
  The third column of Table 4 shows that, when compared toAlexNet ~\cite{krizhevsky01}, VGGNet ~\cite{simonyan02}, and ResNet ~\cite{he04}, the architecture we generated achieved the third highest accuracy of 66.7\%, lower than the accuracy of Highway Net (67.60\%) and ResNet (69.59\%). However, the plot of highest classification accuracy obtained in each generation (see Figure~\ref{fig:acc:f}) shows that the performance of best DCNN architecture we generated has become stable after 10 generations.

  In summary, after evolving ten generations, the proposed genetic DCNN designer generated DCNNs, which performed better than several state-of-the-art DCNNs on three out of six datasets and 
achieved comparable performance on other datasets. It should be note that our genetic DCNN designer is prone to produce DCNNs with less layers.
\section{Discussion on Complexity}
  Although the training of each generated DCNN is limited to 100 epochs, the proposed genetic DCNN designer still has an extremely high computational and space complexity, due to storing and evaluating a large number of DCNN structures. It takes about 3, 5, 18, 12, 11 and 11 GPU-days on average to evolve the DCNN structure for one generation on the MNIST, Fashion-MNIST, EMNIST-Digits, EMNIST-Letters, CIFAR10 and CIFAR100 dataset, respectively. The experiments were conducted using a server with 2 Intel Xeon E5-2678 V3 2.50 GHz and 8 NVIDIA Tesla TiTanXp GPU, 512 GB Memory, 240 GB SSD and Matlab 2017a.
%
\section{Conclusion}
  In this paper, we propose the genetic DCNN designer, an autonomous learning algorithm that can generate DCNN architecture automatically based on the genetic algorithm and data available for the specific image classification problem. This designer is prone to creating DCNNs with less layers. The experimental results on the MNIST, Fashion-MNIST, EMNIST-Digit, EMNIST-Letter, CIFAR10 and CIFAR100 datasets suggest that the proposed algorithm is able to produce automatically DCNN architectures, which are comparative or even outperform the state-of-the-art DCNN models.
%
%
%
%
%

{\small
\bibliographystyle{ieee}
\bibliography{egbib}
}

\end{document}